\documentclass{article}

\usepackage{PRIMEarxiv}

\usepackage[utf8]{inputenc} 
\usepackage[T1]{fontenc}    
\usepackage{hyperref}       
\usepackage{url}            
\usepackage{booktabs}       
\usepackage{amsfonts}       
\usepackage{nicefrac}       
\usepackage{microtype}      
\usepackage{lipsum}
\usepackage{fancyhdr}       
\usepackage{graphicx}       
\graphicspath{{Figures/}}     

\usepackage{color, soul}
\usepackage{graphicx}
\usepackage{float}
\usepackage{makecell}
\usepackage{rotating}
\usepackage{colortbl}
\usepackage{hhline}
\usepackage{caption}
\usepackage{subcaption}

\title{Deep Learning based pipeline for anomaly detection and quality enhancement in industrial binder jetting processes.
\thanks{Presented at conference: 17. Fachtagung "Entwurf komplexer Automatisierungssysteme (EKA)", Magdeburg, June 2022, ISBN: 978-3-948749-23-1}
}

\author{
  Alexander Zeiser \thanks{Corresponding author, contact: \texttt{alexander.az.zeiser@bmw.de}}\\
  Bayerische Motorenwerke AG \\
  Ohmstrasse 2, 84030 Landshut, Germany
   \And
  Bas v. Stein, Thomas Bäck \\
  Leiden Institute of Advanced Computer Science \\
  Niels Bohrweg 1, 2333 CA Leiden, The Netherlands
}

\begin{document}
\maketitle

\begin{abstract}
Anomaly detection describes methods of finding abnormal states, instances or data points that differ from a normal value space. Industrial processes are a domain where predicitve models are needed for finding anomalous data instances for quality enhancement. A main challenge, however, is absence of labels in this environment. This paper contributes to a data-centric way of approaching artificial intelligence in industrial production. With a use case from additive manufacturing for automotive components we present a deep-learning-based image processing pipeline. Additionally, we integrate the concept of domain randomisation and synthetic data in the loop that shows promising results for bridging advances in deep learning and its application to real-world, industrial production processes.
\end{abstract}


\section{Introduction}
Predictive models approach industrial manufacturing complexity by facilitating the assessment of multidimensional relationships. They enable optimisation of process and product quality by unveiling machining dependencies and root-causes of defects. Typically, industrial processes generate heterogeneous datasets consisting of process measurements, quality inspection feedback and maintenance events that only in combination expose causal interactions. One main challenge in industrial production is data quality and pre-processing so that reliable pipelines for modelling can be built. Quickly changing conditions and drift, missing labels, highly imbalanced datasets and noise are some examples of impediments that need to be handled. Due to that, open issues still exist in integrating and exploiting potentials of machine learning in actual series production as surveys and literature on industrial applications reveal \cite{Dogan2021MachineManufacturing}. \newline 
A main challenge we like to highlight in this paper is absence of labels in industrial production data and data processing for in-situ monitoring. This paper contributes to a data-centric way of approaching artificial intelligence in industrial production. With a use case from additive manufacturing for automotive components we present a deep-learning-based image processing pipeline. Additionally, we integrate the concept of domain randomisation and synthetic data in the loop that shows promising results for bridging advances in deep learning and its application to real-world, industrial production processes. \newline
The remainder of this paper is structured as follows. Section \ref{implications} and \ref{anomalyIndustry} introduce  the problem environment, from a broader perspective of data processing and analysis in industrial production to the more focused aspect of anomaly detection for improving process quality. As a use case from production, we present an additive manufacturing process to analyse open issues in the industrial application of an automated process monitoring based on deep learning. Section \ref{pipeline} presents pipeline components and an approach of synthetic data in the loop for developing anomaly detection models with real-world production data. 

\section{Implications of industrial data} \label{implications}
The industrial production system, opposed to a laboratory set-up, is characterised by many environmental and uncontrollable influences, leading to noise and disturbance in data acquisition and mediocre data quality. As a result, any analysis is reliable only after careful data pre-processing \cite{Krau2019SelectionQuality}. Output targets and production key performance indicators, however, pursue high efficiency with low scrap rates and machine down times. Advanced process monitoring as well as predictive methods need to be adaptive and overcome such limitations, like robustness, to be applicable in series production. As stated in \cite{Henke2016THEWORLD} less than 30\% of potential in application of data analytics methods, especially machine learning, is exploited in manufacturing. It illustrates that the gap between advances in research and application in real-world is still high. Highlighted reasons are needed expert knowledge, very problem-specific solutions and uni-dimensional concentration on optimising model output rather than input data.
The importance of a context-based data processing was highlighted also by Andrew Ng with the 'Data-Centric AI Competition' and adopted by Motamedi et al. \cite{Motamedi2021AData}, focusing on actions of dataset preparation and data quality enhancement before training and fine-tuning a predictive model (data-centric before model-centric). Ng shows that an increase in data quality can exert far greater influence on the prediction accuracy than hyperparameter optimisation of a machine learning model alone \cite{Ng2021MLOps:AI}. In the context of real-world industrial applications this is an encouraging and advisable approach as well to develop maturity levels further (from descriptive towards self-optimising/prescriptive). Main success factor for achieving promising results is representative, sufficiently large and high-quality data.  \newline
Another aspect of industrial processes is the amount of data it creates and is available for knowledge discovery. However, the process from merging of data bases to actual analysis becomes a difficult task if unique identifiers, timestamps and labels (e.g. from quality inspection) are missing. Unsupervised approaches are favourable in these environments to get insights from data, nevertheless. Another approach is domain randomisation or adaptation, hence, utilising synthetic data, either simulated or abstracted from the real process. Major benefits are creation of labelled data and the high amount of data at low cost, especially for under-represented classes as in anomaly detection cases. In principle, the concept intends to reach high generalisation for the real-world data by transfer learning from synthetic data. \cite{Tobin2017DomainWorld, Valtchev2021DomainClassification}

\section{Anomaly detection in industrial processes} \label{anomalyIndustry}
Anomaly detection describes methods of finding abnormal states, instances or data points that differ from a normal value space \cite{Goldstein2016AData}. The three categories of supervised, semi-supervised and unsupervised each summarise different techniques and methods from statistics (e.g. z-score) or machine learning (e.g. One-Class SVM, Autoencoders, LSTM). The term is predominantly used for highly unbalanced problems. Often no labels are available for learning or classes for different states are not known \cite{Chandola2012AnomalySurvey}. The unsupervised approach is applied in several domains, such as medicine (e.g. detection of critical cardiac arrhythmia, tumor detection with computed tomography), banking (e.g. fraudulent financial transactions, payments with stolen credit cards), security (e.g. surveillance, document forgery, network intrusion) but also engineering (e.g. critical state detection) \cite{Goldstein2016AData}. Besides point anomalies (one data instance lies out of the normal data region) Chandola et al.~define two other types of anomalies: contextual anomalies (instance of data is anomalous only in a certain context but not in another) and collective anomalies (individual values lie within normal data region but as a collection of related data instances they form an anomaly) \cite{Chandola2009AnomalySurvey}. 

\subsection{Use case: Binder jetting additive manufacturing} \label{usecase}
Additive manufacturing (AM) is a widely used technology that comprises many sub-technologies. What most have in common is a layer-wise construction with deposition of new material. However, the actual way of how these layers are created differentiates sub-technologies and is standardised by ISO 17296 \cite{ISO/TC2612015ISOFeedstock}.
One major benefit over classic manufacturing techniques is free form design of complex geometries. Another advantage is a tool-independent production of different shapes and therefore a quick adaptation to new designs or product updates. Typically, AM is used in rapid prototyping, small series and small dimensions. However, advances in technology enable also medium to high scale production of specific components. Also larger components can be manufactured with AM technologies, especially binder jetting. Due to comparably low working temperature nearly no heat induced shrinkage, cracks or porosities appear that typically hinder other AM technologies from manufacturing bigger dimensions \cite{Gibson2021AdditiveTechnologies}.
In series production binder jetting is often combined in a multi-stage process with other manufacturing techniques, like casting. One example are automotive cylinder heads of BMW straight-four and straight-six engines. Advantages of both conventional and modern manufacturing are combined in this way. In the following current developments are described.

\subsection{Related work}
A majority of research on optical process monitoring and computer vision related to additive manufacturing is concentrating on powder bed fusion. Quality and defect predictions are built especially on (image) data from the melt pool. Nevertheless, also other AM processes are adduced for image-based process monitoring and anomaly detection for quality enhancement. A comprehensive summary can be found in  \cite{Gierson2021MachineManufacturing}. 

With a focus on image processing, various algorithms are discussed in \cite{Trinks2019SmartMining}. In an application to a small sample of image data from an extrusion process they are compared based on accuracy for defect prediction. The trained data model is constantly fed with a stream of new images and classifies in real time. As soon as a production error is discovered, the process is automatically stopped. However, issues like generalisation and processing of large datasets hinder application in series production. Günther et al.~\cite{Gunther2020ConditionApproaches} describe requirements for condition monitoring for binder jetting and propose an image-based defect detection. Research is focused on nozzle failures that lead to work-piece defects. In a series of steps the  work-piece shape is extracted from the recorded image and defect analysis is performed based on the transformed binary image. The distribution density of black and white pixels along the printing direction indicate a printing failure \cite{Gunther2020ConditionApproaches}.

Current literature presents and discusses  approaches for process monitoring and defect detection with computer vision. Even though the need for anomaly detection in processes with un-labelled data is stated research concentrates on process engineering parameter setting and is often related to detection of material-specific and mechanical defects \cite{Scime2020Layer-wiseProcesses, Gierson2021MachineManufacturing}. The development of methods for bigger datasets, series production and the implication associated with it is still at an early stage.

\section{Data processing pipeline for the AM binder jetting in-situ monitoring} \label{pipeline}
In order to bring anomaly detection into deployment, first, process historic data is analysed and used for development. The pipeline components are described in the following.

\subsection{Data acquisition}
\begin{figure}[H] \vspace{4pt}
    \centering
    \captionsetup{justification=centering}
    \noindent\hbox to 0.52\textwidth{
    \begin{subfigure}[t]{0.38\textwidth}
        \includegraphics[width=\textwidth]{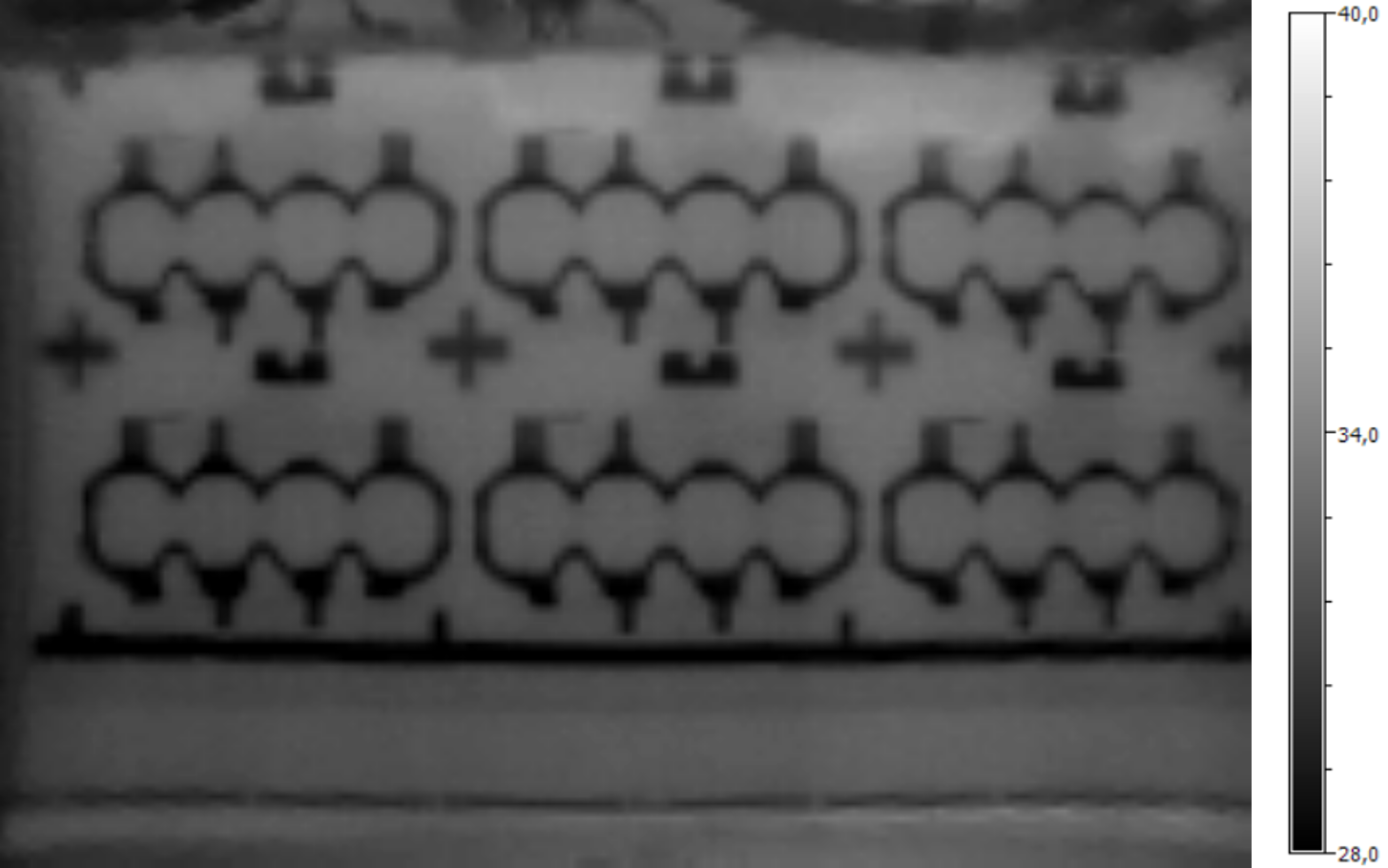}
        \caption{Original thermal image in grayscale}
        \label{fig:ImgGray}
    \end{subfigure}
    }
    \begin{subfigure}[t]{0.38\textwidth}
        \includegraphics[width=\textwidth]{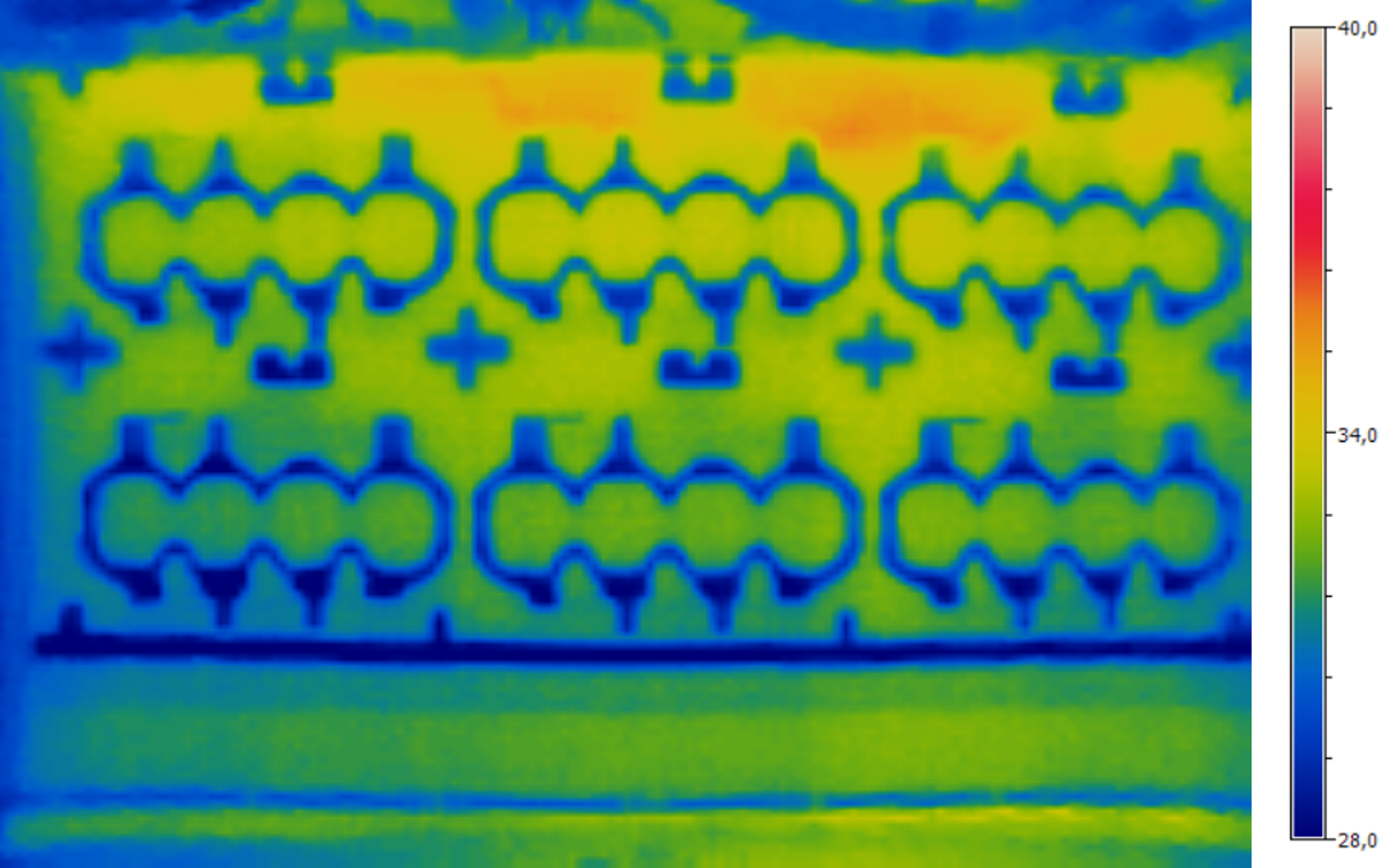}
        \caption{False colour image for improved temperature visualisation}
        \label{fig:ImgCol}
    \end{subfigure}
    \caption{Example thermal image from print job (layer 500/700)}
    \label{fig:ExmpImg}
    \vspace{-8pt} 
\end{figure}

Data from manufacturing at BMW plant Landshut consists of processing measurements, machining parameters, ambient conditions and quality inspection. However, in focus of our work is image data coming from an infrared (IR) camera mounted in the inside of the printing room. During the process of sand and binder application an infrared light activates the binder that bonds loose sand particles. 
With a normal camera setup only a flat sandy surface can be seen. However, as the process heats up the whole powder bed, shapes of the printed part become visual by infrared imaging. Another benefit of IR-imaging by visualising temperatures is to make information of energy deposition onto the print bed available for process monitoring. Since binder and energy deposition have a major influence on dimensional accuracy a systematic image data analysis is needed to support process optimisation decisively. In terms of a live monitoring system the objective is to detect temperature and geometric anomalies and provide information to a worker about location, layer and severity. The setup in place is triggered automatically by the machine control unit (PLC) and generates an image per each layer. Depending on the part produced a complete image stack consists of 600-800 images. An example is shown in Fig. \ref{fig:ExmpImg}. Images are recorded in grayscale where the pixel values (0-black, 255-white) correspond to a fixed range of temperatures to assure comparability between print jobs and over time. 

\subsection{Data cleaning}

Data cleaning is performed to have a dataset free of incorrectly acquired records and an important step for improving data quality. With the goal of learning patterns it is necessary, especially for training machine learning models, to not detract learning from erroneous data that falsifies underlying dependencies. However, at the same time the cleaned dataset must still  accurately represent the distribution of the real sample to generalise well \cite{Lee2021APerformance}.
\begin{figure}[H] \vspace{4pt}
    \centering
    \includegraphics[scale=0.4]{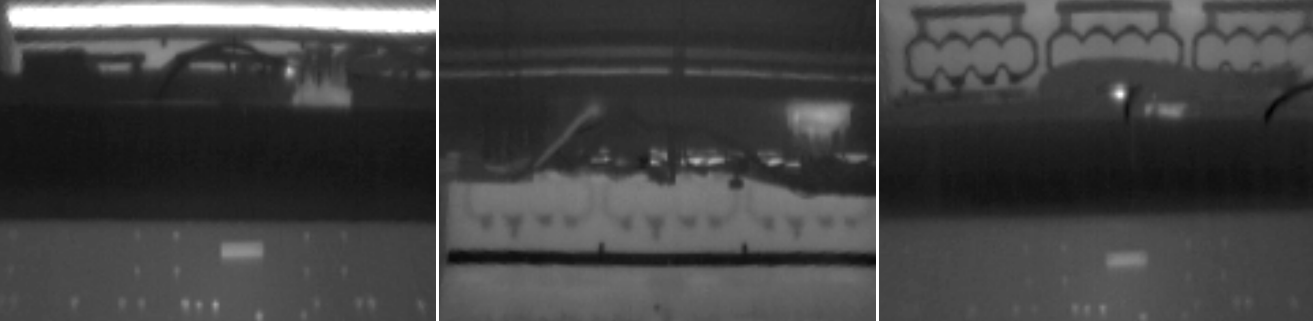}
    \caption{Example of Type 1 anomalies (hidden print bed)}
    \label{fig:badImg}
    \vspace{-8pt}    
\end{figure}

\begin{figure}[H] \vspace{4pt}
    \centering
    \captionsetup{justification=centering}
    \includegraphics[scale=0.6]{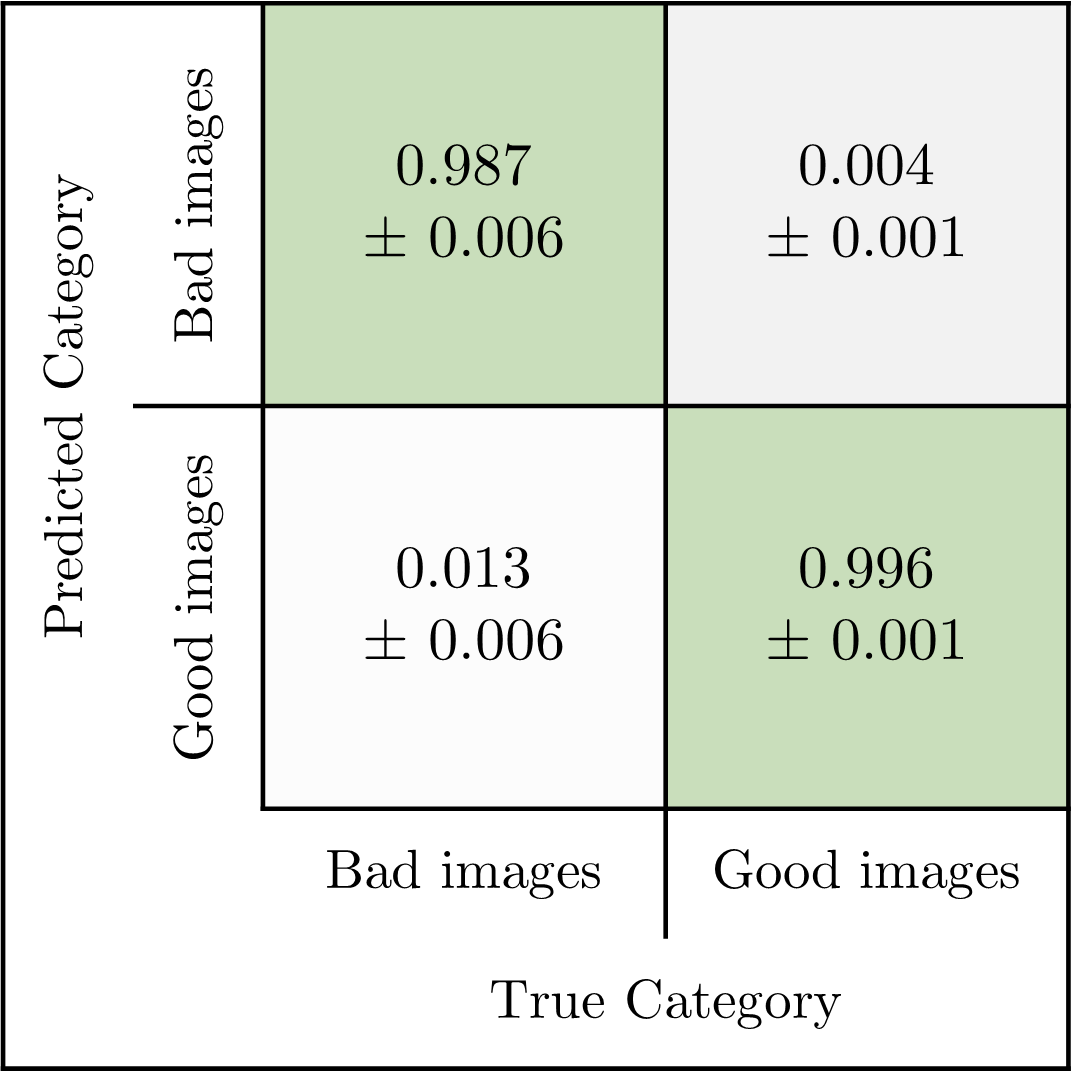}
    \caption{Normalised Confusion Chart for test set (1000 images): \\ mean and confidence interval of 10 repetitions (F1 score: 0.98)}
    \label{fig:confChart}
    \vspace{-8pt} 
\end{figure}

For the binder jetting process, two types of anomalies must be distinguished. First type (Type 1 anomaly) is related to possible sensor malfunction or human intervention that leads to incorrect data acquisition. These anomalies can also be referred to as actual process flow outliers that lead to data quality issues. Examples are shown in Fig.~\ref{fig:badImg}, where the camera triggered in a wrong process instance. The anomalous or outlier image, in this context, does not present a clear view onto the print bed.
Type 2 anomalies, in contrast, are defined as deviations from the normal manufacturing procedure, potentially resulting in product related quality issues like defects or dimensional inaccuracy. Data cleaning refers only to Type 1 anomalies since Type 2 anomalies are the actual matter of investigation in the dataset. In later model development for Type 2 anomaly detection the Type 1 anomalies must be filtered out to minimise pseudo-defects. Since it is a classification task a deep learning model can potentially solve the distinction of good and bad images for Type 1 anomalies. This has been tested with a convolutional neural network (CNN) of six convolutional 2D layers, each followed by 2D max-pooling and batch normalisation layer. Finally, a flatten layer and two dense layers are used for classification (activation function: ReLU, optimiser: Adam with default values). Early stopping is applied based on validation accuracy. The training set consists of  22000 images (70/30 \% good/bad), of which 20\% are hold out for validation. Train and test images have been randomly sampled from different jobs and printers over a production period of 3 months. 10 learning repetitions were carried out with randomised weights initialisation. Standard data augmentation was performed on the training data within the Tensorflow model. Testing was performed with another 5000 images (70/30 \% good/bad distribution) on each of the 10 models.
The mean performance including 95\%-confidence interval is shown in Fig. \ref{fig:confChart} in terms of the confusion chart. The network is clearly able to distinguish well between 'good' images of the process working fine and 'bad' images (Type 1 anomalies) as exemplified in Fig. \ref{fig:badImg}.
Labelling was performed manually. It must be pointed out that for process flow outlier images this is still a bearable effort since these Type 1 anomalies are simple to distinguish for the human vision. Type 2 anomalies must be treated differently as defect characteristics may not that explicit.

\subsection{Data preparation}
Due to space constraints, wide-angle lenses with 80° opening angles are used that create a radial distortion. By printing straight checkerboard patterns, this lens distortion can be calibrated. New images are undistorted by the camera matrix and distortion coefficients in a post-processing step. Camera calibration functions of the Python package OpenCV are implemented \cite{OpenCV2021OpenCVCalibration}. Checkerboards of different dimensions and viewing angles are tested and examined by the re-projection error.
As multiple machines work in parallel, slightly distinct camera mounting positions result in different viewing angles onto the print bed. To align images an image registration algorithm is implemented using OpenCV. Here, the enhanced correlation coefficient maximisation (ECC) algorithm, adopted from \cite{Evangelidis2008ParametricMaximization}, outperforms other feature-based algorithms like Oriented FAST and Rotated BRIEF (ORB) with brute force matching. ECC is independent from photo-metric distortions like contrast and brightness, hence, well applicable to thermal images where temperatures actually translate into brightness or colour. Transformation parameters can be calculated by target and source image of the exact same layer and is valid for the whole stack of images of one print job. Finally, images are cropped to the region of interest, namely the print bed.

\subsection{Synthetic data in the loop for model development} 

The aim of a live monitoring system is to detect anomalies in (near-)real time  that become sources of product defects. This information must be passed to a worker to either stop the print job, to scrap affected parts directly after process completion without quality test or to perform a more target-oriented inspection. Additionally, process optimisation by parameter adaptation is a prospective way to approach quality enhancement in the long term.

\begin{figure}[H] \vspace{4pt}
    \centering
    \captionsetup{justification=centering}
    \includegraphics[scale=0.8]{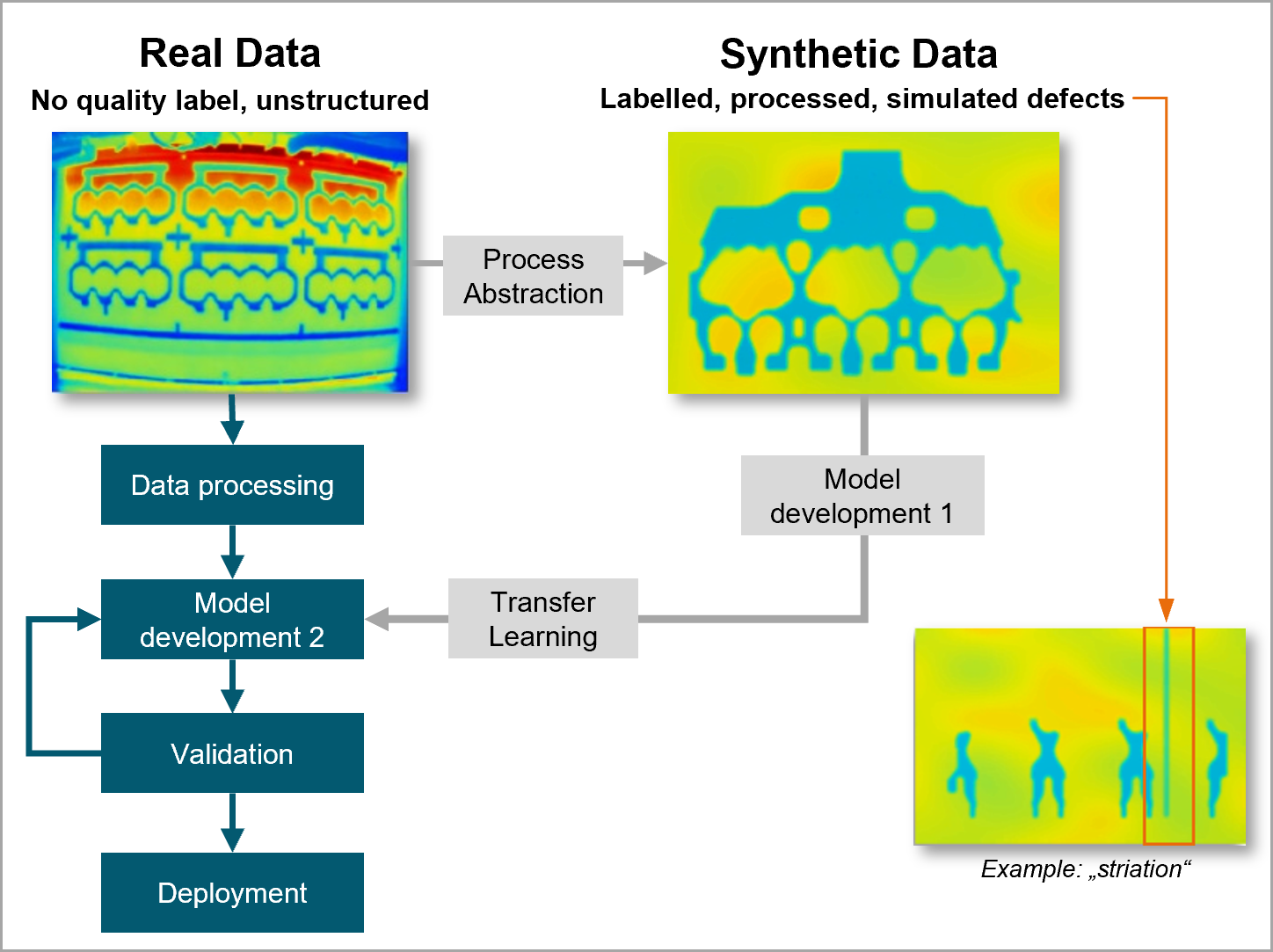}
    \caption{Synthetic data creation by process abstraction}
    \label{fig:transfLearn}
    \vspace{-8pt} 
\end{figure}

\begin{figure}[H] \vspace{4pt}
    \centering
    \captionsetup{justification=centering}
    \noindent\hbox to 0.52\textwidth{
    \begin{subfigure}[t]{0.4\textwidth}
        \includegraphics[width=\textwidth]{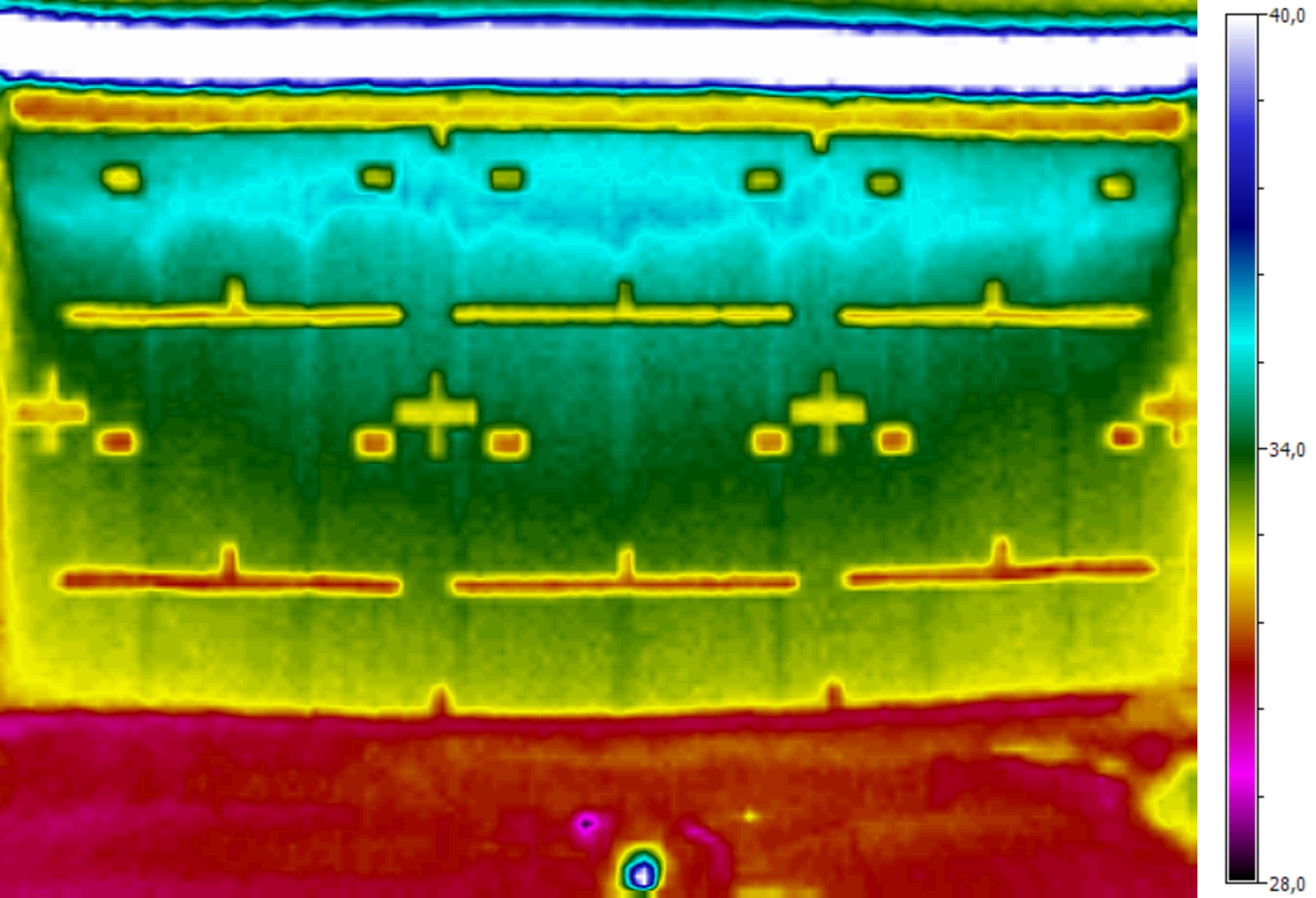}
        \caption{Colour-manipulated image for pattern emphasis (layer 75/700)}
        \label{fig:ManipImg}
    \end{subfigure}
    }
    \begin{subfigure}[t]{0.4\textwidth}
        \includegraphics[width=\textwidth]{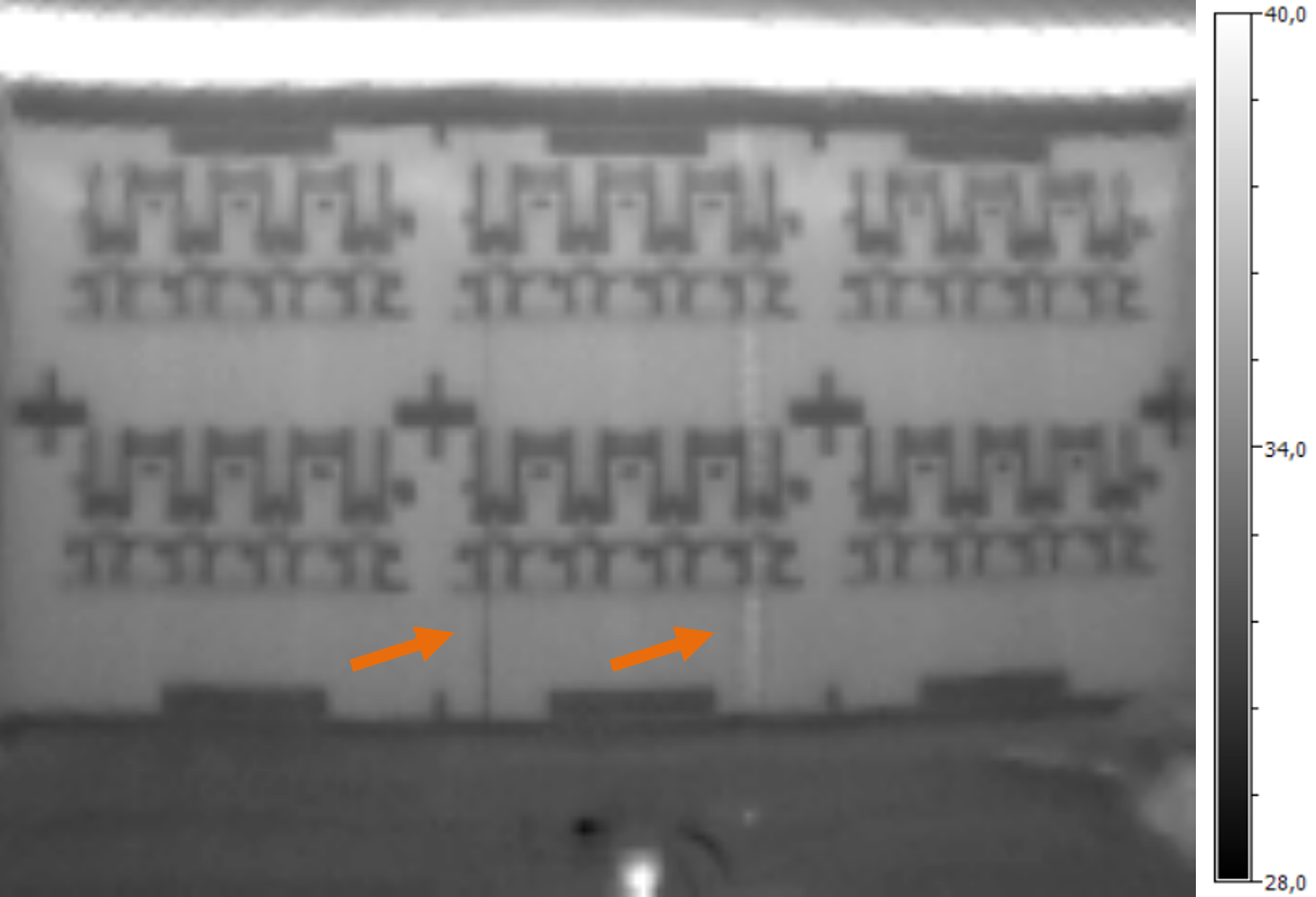}
        \caption{Striation over whole print bed (layer 400/700)}
        \label{fig:Abzieher}
    \end{subfigure}
    \caption{Type 2 anomalies.}
    \vspace{-8pt} 
\end{figure}

Fig. \ref{fig:Abzieher} shows a noticeable Type 2 anomaly, namely striations. Vertical striations from excess binder, sand or nozzle-clogging, foreign objects and porosities can lead to work-piece internal defects that remain hidden during visual inspection. Resulting layer disintegration that affects the material strength cause broken sand cores during subsequent casting steps. Other defects like layer shifts and sand agglomerates cause dimensional inaccuracies. Due to the large amount of image data in the industrial setup a manual labelling and training set preparation is very cost intensive. We therefore see huge potential in approaching the anomaly detection problem with transfer learning from synthetic data that can be generated and automatically labelled in only a fraction of time. In the AM binder jetting case this data is abstracted from the real process. Fig. \ref{fig:transfLearn} illustrates the workflow of integrating synthetic data into model development for the real data. It follows the hypothesis that learning from synthetic data allows for generalisation to real-world data. We utilise the Standard Tessellation Language (stl) file generated from the work-piece CAD file. In the same way as preparing the AM print job, we slice the stl file into layers and create Support Vector Graphic (svg) files that define the layerwise work-piece contours. We adapt domain randomisation in terms of background and work-piece grayscale dispersion, work-piece position, layer-to-layer translucency and blurring of work-piece contours. 
\begin{figure}[H] \vspace{4pt}
    \centering
    \captionsetup{justification=centering}
    \includegraphics[scale=0.35]{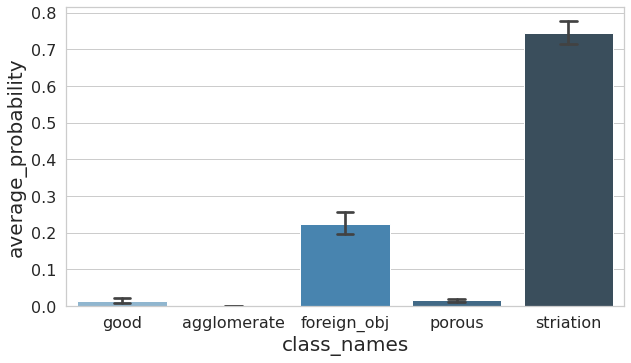}
    \caption{Predicted class membership probability for real striation images}
    \label{fig:ClassStriation}
    \vspace{-8pt} 
\end{figure}

This procedure is performed layerwise as well as jobwise to increase variability. The randomisation intervals are defined by min and max values of a random subsample of real images. Within this procedure a set of aforementioned defects are defined visually and placed at certain layers. Synthetic images are labelled correspondingly. With the resulting dataset of clean and anomalous jobs a neural network can be trained.

We use similar CNN architecture and parameters as for the Type 1 anomaly classification, with a softmax layer added for prediction of class membership probability. Train data set consists of 12500 synthetic images, each 2500 per class (good/no defect, agglomerates, foreign objects, porous and striation). Fig. \ref{fig:ClassStriation} shows test results for a set of 680 images of class 'striation'. On average, the model decided with confidence of 74\% that images show a 'striation'. Followed by defect type 'foreign object' with 22\% probability. The predicted probability for the images to fall within the 'good' class is at only less than 2\% on average. 

\section{Conclusion}
One major challenge in advancing deep learning potentials for industrial processes, e.g. quality improvements, is access to clean and labelled data for model development. A more data-centric approach to artificial intelligence shifts attention towards data pre-processing and quality. With this input optimisation deep learning, like anomaly detection techniques will likely perform better in real-world scenarios. We presented a processing pipeline for in-situ monitoring of an AM binder jetting process. Bringing synthetic data into the loop for development of this pipeline shows great potential. Economically due to low cost of creation without any physical production, as well as technically due to exact labelling. First results on defect prediction with a CNN trained only by synthetic data are promising to investigate this approach further. Future work will direct towards enhancing domain randomisation in order to explore effects of variability levels in synthetic data. Combining synthetic data with real data in a second modelling step is expected to increase performance. \newline
At this stage, tests were only performed with defect type striation. In the future we will extend tests to other defects. Also, only object-like defects were considered as anomalies so far. Additionally, trends, patterns and higher order features may be present as exemplified by Fig. \ref{fig:ManipImg}. Here, several patterns become noticeable only after manipulation. Deeper analysis on unknown, more implicit anomalies must be performed. Autoencoders present high potential for the kind of task. Generative models are also further to be investigated due to their semi-supervised learning capabilities, combining unlabelled big data with labelled small data or using combinations of real and synthetic data. Another aspect where research will be focused on is time and spatial dependency of (thermal) images as input for anomaly detection. Approaches of combining different neural network types like CNN with Recurrent Neural Networks (e.g. LSTM) to include the aspect of time dependency are discussed in recent literature but need further elaboration for a series production case \cite{Zuo2015ConvolutionalRepresentation}.

\bibliographystyle{unsrt}  
\bibliography{MAIN}

\begin{thebibliography}{10}

\bibitem{Dogan2021MachineManufacturing}
Alican Dogan and Derya Birant.
\newblock {Machine learning and data mining in manufacturing}.
\newblock {\em Expert Systems with Applications}, 166:114060, 2021.

\bibitem{Krau2019SelectionQuality}
Jonathan Krau{\ss}, Maik Frye, Gustavo Teodoro, Döhler Beck, and Robert~H
  Schmitt.
\newblock {Selection and Application of Machine Learning- Algorithms in
  Production Quality}.
\newblock In Jürgen Beyerer, editor, {\em International Conference ML4CPS},
  pages 46--57, 2019.

\bibitem{Henke2016THEWORLD}
Nicolaus Henke, Jacques Bughin, and Michael Chui.
\newblock {The Age of Analytics: Competing in a Data-Driven World}.
\newblock {\em McKinsey {\&} Company}, 2016.

\bibitem{Motamedi2021AData}
Mohammad Motamedi, Nikolay Sakharnykh, and Tim Kaldewey.
\newblock {A Data-Centric Approach for Training Deep Neural Networks with Less
  Data}.
\newblock In {\em 35th Conference on Neural Information Processing Systems
  (NeurIPS 2021)}, 10 2021.

\bibitem{Ng2021MLOps:AI}
Andrew Ng.
\newblock {MLOps: From Model-centric to Data-centric AI}.
\newblock
  https://www.deeplearning.ai/wp-content/uploads/2021/06/MLOps-From-Model-centric-to-Data-centric-AI.pdf,
  2021.

\bibitem{Tobin2017DomainWorld}
Josh Tobin, Rachel Fong, and Alex Ray.
\newblock {Domain randomization for transferring deep neural networks from
  simulation to the real world}.
\newblock In {\em 2017 IEEE/RSJ International Conference on Intelligent Robots
  and Systems (IROS)}, pages 23--30. IEEE, 9 2017.

\bibitem{Valtchev2021DomainClassification}
Svetozar~Zarko Valtchev and Jianhong Wu.
\newblock {Domain randomization for neural network classification}.
\newblock {\em Journal of Big Data}, 8(1):94, 12 2021.

\bibitem{Goldstein2016AData}
Markus Goldstein and Seiichi Uchida.
\newblock {A comparative evaluation of unsupervised anomaly detection
  algorithms for multivariate data}.
\newblock {\em PLoS ONE}, 11(4), 2016.

\bibitem{Chandola2012AnomalySurvey}
Varun Chandola, Arindam Banerjee, and Vipin Kumar.
\newblock {Anomaly detection for discrete sequences: A survey}.
\newblock {\em IEEE Transactions on Knowledge and Data Engineering},
  24(5):823--839, 2012.

\bibitem{Chandola2009AnomalySurvey}
V~Chandola, A~Banerjee, and V~Kumar.
\newblock {Anomaly detection: A survey}.
\newblock {\em ACM Reference Format}, 41(15), 2009.

\bibitem{ISO/TC2612015ISOFeedstock}
{ISO/TC 261}.
\newblock {ISO 17296-2:2015 Additive manufacturing — General principles —
  Part 2: Overview of process categories and feedstock}, 2015.

\bibitem{Gibson2021AdditiveTechnologies}
Ian Gibson, David~W. Rosen, and Brent Stucker.
\newblock {\em {Additive Manufacturing Technologies}}.
\newblock Springer US, Boston, MA, 2021.

\bibitem{Gierson2021MachineManufacturing}
Dean Gierson and Allan Rennie.
\newblock {Machine Learning for Advanced Additive Manufacturing}.
\newblock {\em Encyclopedia}, 3(5):576--588, 2021.

\bibitem{Trinks2019SmartMining}
Sebastian Trinks and Carsten Felden.
\newblock {Smart Factory – Konzeption und Prototyp zum Image Mining}.
\newblock {\em HMD Praxis der Wirtschaftsinformatik}, 56(5):1017--1040, 10
  2019.

\bibitem{Gunther2020ConditionApproaches}
Daniel G{\"{u}}nther, Mina Fahimi~Pirehgalin, Iris Wei{\ss}, and Birgit
  Vogel-Heuser.
\newblock {Condition monitoring for the Binder Jetting AM-process with machine
  learning approaches}.
\newblock In {\em 2020 IEEE Conference on Industrial Cyberphysical Systems
  (ICPS)}, volume~1, 2020.

\bibitem{Scime2020Layer-wiseProcesses}
Luke Scime, Derek Siddel, and Seth Baird.
\newblock {Layer-wise anomaly detection and classification for powder bed
  additive manufacturing processes}.
\newblock {\em Additive Manufacturing}, 36(March):101453, 12 2020.

\bibitem{Lee2021APerformance}
Ga~Young Lee, Lubna Alzamil, Bakhtiyar Doskenov, and Arash Termehchy.
\newblock {A Survey on Data Cleaning Methods for Improved Machine Learning
  Model Performance}.
\newblock 2021.

\bibitem{OpenCV2021OpenCVCalibration}
{OpenCV}.
\newblock {OpenCV Camera Calibration}.
\newblock {\em https://docs.opencv.org/4.x/dc/dbb/
  tutorial{\_}py{\_}calibration.html}, 2021.

\bibitem{Evangelidis2008ParametricMaximization}
G.D. Evangelidis and E.Z. Psarakis.
\newblock {Parametric Image Alignment Using Enhanced Correlation Coefficient
  Maximization}.
\newblock {\em IEEE Transactions on Pattern Analysis and Machine Intelligence},
  30(10):1858--1865, 10 2008.

\bibitem{Zuo2015ConvolutionalRepresentation}
Zhen Zuo, Bing Shuai, and Gang Wang.
\newblock {Convolutional recurrent neural networks: Learning spatial
  dependencies for image representation}.
\newblock In {\em IEEE Computer Society Conference on Computer Vision and
  Pattern Recognition Workshops}, volume 2015-Octob, pages 18--26, 2015.

\end{thebibliography}

\end{document}